\documentclass[manuscript,screen]{acmart}

\AtBeginDocument{%
  \providecommand\BibTeX{{%
    \normalfont B\kern-0.5em{\scshape i\kern-0.25em b}\kern-0.8em\TeX}}}

\setcopyright{acmcopyright}
\copyrightyear{2023}
\acmYear{2023}
\acmDOI{XXXXXXX.XXXXXXX}

\acmConference[TBD]{}{}{}
\acmPrice{15.00}
\acmISBN{978-1-4503-XXXX-X/18/06}

\usepackage{xspace}
\usepackage{bbm}

\newcommand{\iid}{i.i.d.\xspace}
\newcommand{\ciid}{conditional-i.i.d.\xspace}
\newcommand{\Ciid}{Conditional-i.i.d.\xspace}
\newcommand{\ie}{i.e.,\xspace}
\newcommand{\eg}{e.g.,\xspace}

\begin{document}

\title{The Unbearable Weight of Massive Privilege: Revisiting Bias-Variance Trade-Offs in the Context of Fair Prediction}


\author{Falaah Arif Khan}
\affiliation{%
  \institution{New York University}
  \city{New York}
  \country{USA}}
\email{fa2161@nyu.edu}

\author{Julia Stoyanovich}
\affiliation{%
  \institution{New York University}
  \city{New York}
  \country{USA}
}

\renewcommand{\shortauthors}{Arif Khan, et al}

\begin{abstract}
In this paper we revisit the bias-variance decomposition of model error from the perspective of designing a fair classifier: we are motivated by the widely held socio-technical belief that noise variance in large datasets in social domains tracks demographic characteristics such as gender, race, disability, etc. We propose a \ciid model built from group-specific classifiers that seeks to improve on the trade-offs made by a single model (\iid setting). We theoretically analyze the bias-variance decomposition of different models in the Gaussian Mixture Model, and then empirically test our setup on the COMPAS and folktables datasets. We instantiate the \ciid model with two procedures that improve ``fairness'' by conditioning out undesirable effects: first, by conditioning directly on sensitive attributes, and second, by clustering samples into groups and conditioning on cluster membership (blind to protected group membership). 

Our analysis suggests that there might be principled procedures and concrete real-world use cases under which conditional models are preferred, and our striking empirical results strongly indicate that non-\iid settings, such as the \ciid setting proposed here, might be more suitable for big data applications in social contexts.
\end{abstract}


\maketitle

\section{Motivation}
\label{sec:intro}

In this paper we challenge the dominant modelling paradigm in fair machine learning, namely, that data samples are \emph{independent and identically distributed} (\iid) from an underlying data generating process. Instead, we propose to construct fair estimators under \ciid assumptions. 
As the name suggests, \ciid challenges the assumption  that all samples come from the same distribution (\ie that they are identically distributed).  


\begin{figure*}[h!]
    \centering
    \includegraphics[width=\linewidth]{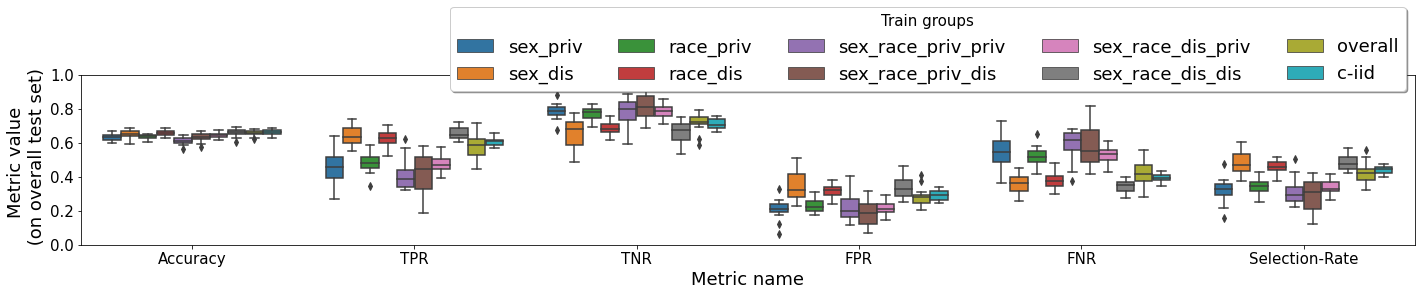}
 \vspace{-0.75cm}
    \caption{Conditioning on sensitive attributes, COMPAS: Test performance of different models on the overall test set.}
    \label{fig:compas_all}
\end{figure*}

\begin{figure*}[h!]
    \centering
    \includegraphics[width=\linewidth]{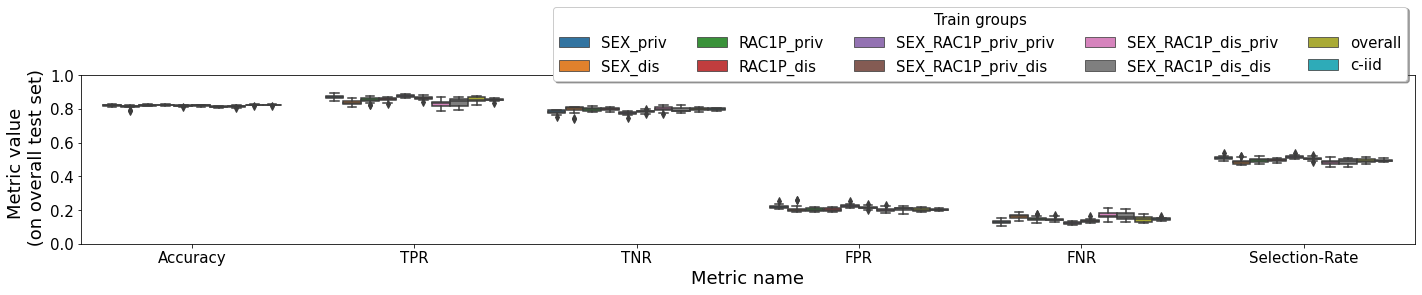}
     \vspace{-0.75cm}
    \caption{Conditioning on sensitive attributes, folktables: Test performance of different models on the overall test set.}
    \label{fig:folk_all}
\end{figure*}

We are motivated by the widely-held sociological view 
that the data from marginalized social and demographic groups tends to be ``noisier'' than the data from privileged social groups~\cite{kappelhof2017,schelter2019fairprep}. 
In alignment with the \emph{WAE} worldview~\cite{friedler_impossibility}, we posit that, while people's 
qualifications are in fact drawn from the same distribution (equal natural talent), social inequality affects different demographic groups differently (unequal treatment). 
The composite effect of equal natural talent and unequal treatment results in distributions that are no longer identical.  
The intuition is then that, since different social groups are assumed to be drawn from different distributions, we can improve on the bias-variance trade-offs of a single model that treats the data as \iid by training multiple group-specific conditional models.   

In this paper, we formally and empirically study the bias-variance trade-off under \ciid.  Notably, it may be impractical --- or even illegal --- to train models for different demographic groups and then select a model to use at decision time based on an individual's group membership.\footnote{In the U.S. this is illegal in many domains based on the doctrine of disparate treatment, see \url{https://en.wikipedia.org/wiki/Disparate_treatment}.}  We show that understanding the bias-variance trade-off under \ciid helps select a specific \emph{single model} to use at decision time,  and, furthermore, that the best-performing model is not always the one that is trained under \iid (\ie on the entire training dataset).   Our results are highlighted in Figures~\ref{fig:compas_all} and~\ref{fig:folk_all}: Group-specific models that are only trained on samples from intersectionally disadvantaged groups, which form 17.7$\%$ of the folktables dataset~\cite{DBLP:conf/nips/DingHMS21} and 49.1$\%$ of the COMPAS dataset~\cite{compas_propublica}, respectively, are highly competitive on the overall test set, and outperform \iid models on disadvantaged test groups.
 


{\bf In this paper, we make the following contributions:}
 
(1) We propose a new modelling paradigm for fair machine learning, namely, the \ciid (Section~\ref{sec:ciid}).\\
(2) We theoretically contrast the bias-variance trade-offs made in the dominant \iid setting and in the new \ciid setting using the example of mean estimation in the Gaussian Mixture Model (Section \ref{sec:theory}). We find that the \iid and \ciid models lie on opposite ends of bias-variance spectrum: the \iid model has low variance, but potentially unbounded bias in poorly specified settings, whereas the \ciid model is unbiased (on the demographic group it saw during training) by construction, but it has large variance. However, in some decision-making contexts --- specially in critical social contexts --- sacrificing estimator variance to gain unbiasedness on protected groups of interest is a worthy trade-off.\\
(3) We empirically evaluate our approach on two real-world benchmarks, and show that \ciid models are competitive with \iid models (Section \ref{sec:experiments}). We  demonstrate two procedures that improve ``fairness'': (a) conditioning on the sensitive attributes and (b) clustering samples into groups and conditioning on cluster membership. 

\section{The Conditional-i.i.d Model}
\label{sec:ciid}
Assume we are given a dataset of covariates and targets $\mathcal{D}(X^\text{i}, Y^\text{i})$, and we wish to learn an estimator $\hat{f} = E[Y|X=x]$, which takes the covariates of an unseen sample $x$, and returns its predicted label $\hat{y} = \hat{f}(x)$. Further, assume that the covariate vector $X^i$ can be partitioned into two types of features: $X_{relevant}$ and $X_{protected}$, and that demographic groups are constructed on the basis of $X_{protected}$. 
Let $x^*$ 
be the value of the feature $X_{protected}$ for samples from the privileged group.  Then:
$$ \mathcal{D}^{priv} = \{ (X^i,Y^i)| X^i_{protected} = x^* \}, i=1,2 \dots n $$
$$ \mathcal{D}^{dis} = \{ (X^i,Y_i) | X^i_{protected} \neq x^* \}, i=1,2 \dots n $$
$$ \mathcal{D} = \mathcal{D}^{priv} \cup \mathcal{D}^{dis}$$

Conventionally, we assume that samples $(X_i,Y_i)$ are \iid: 
$$\mathcal{D}, \mathcal{D}^{priv}, \mathcal{D}^{dis} \sim (\mathcal{X},\mathcal{Y})$$
Instead, in this paper we model the \ciid setting:
$$\mathcal{D}^{priv} \sim (\mathcal{X}^{priv},\mathcal{Y}^{priv}), \mathcal{D}^{dis} \sim (\mathcal{X}^{dis},\mathcal{Y}^{dis})$$
Under \ciid, the \iid assumption holds after conditioning on the protected attributes.

Different statistical measures correspond to different notions of fairness, but a particularly influential approach has been to 
assume all samples are \iid, fit an estimator to the full dataset (all groups), and then post-process predictions until the desired group-wise parity (fairness) constraint is met \cite{hardt_EOP2016}. The main observation we make here is that there is a global trade-off that the overall (iid) estimator $\hat{Y}$ is making. 
The goal of fair-ML is to find a principled way to trade off comparable performance within groups with slightly worse performance overall. 
%
In this paper, we interrogate this procedural tension: 
adopting \ciid assumptions, we train group-specific models and investigate the trade-offs they make in terms of statistical bias and variance compared to the single (\iid) model. 


\section{Gaussian Mixture Model Mean Estimation}
\label{sec:theory}

\begin{table*}[t!]
\centering 
\small 
\caption{Summary of bias-variance trade-offs of different mean estimators}
\centering
\begin{tabular}{||c |c |c |c ||}
\hline
Model & Bias on $priv$ & Bias on $dis$ & Variance  \\ [0.5ex]
\hline\hline
overall (\iid) & $p_{dis}.\mathbb{E}[\Delta \mu]$ & $p_{priv}.\mathbb{E}[\Delta \mu]$ & $\frac{1}{n} [\frac{n_{priv}}{n}.\sigma^2_{priv} + \frac{n_{dis}}{n}.\sigma^2_{dis}]$ \\
\hline
ensemble & $\frac{1}{2} \mathbb{E} [\Delta \mu]$ & $\frac{1}{2} \mathbb{E} [\Delta \mu]$ & $\frac{1}{4} (\frac{\sigma^2_{priv}}{n_{priv}} + \frac{\sigma^2_{dis}}{n_{dis}})$ \\
\hline
disprivileged & $\mathbb{E} [\Delta \mu]$ & 0 & $\sigma^2_{dis}/n_{dis}$\\
\hline
\ciid & 0 & 0 & $\sigma^2_{priv}/n_{priv}$ (on priv), $\sigma^2_{dis}/n_{dis}$ (on dis) \\

\hline
\end{tabular}
\label{table:mean-results}
\end{table*}

A natural way to think about the \ciid setting is through the Gaussian mixture model (GMM), where the latent variable that determines which component of the mixture the sample is drawn from is no longer latent/unobserved, but is one of the covariates, specifically $X_{protected}$. We will now use mean estimation in GMM, to build intuition about the bias-variance trade-offs under \ciid. 
\begin{figure}[b!]
    \centering
    \includegraphics[width=8cm]{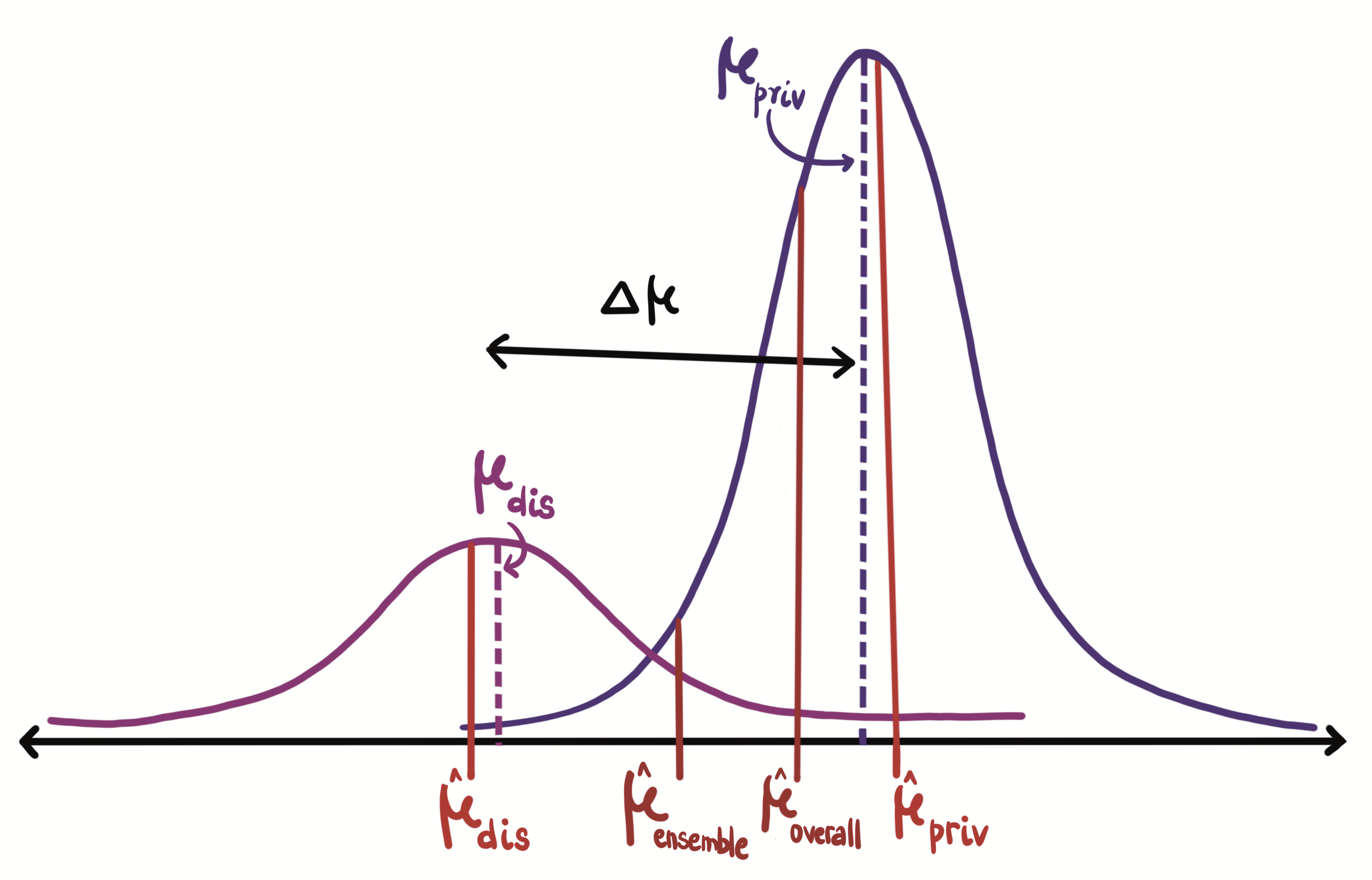}
    \caption{Mean estimation under \ciid.}
    \label{fig:mean_estimation}
\end{figure}
\subsection{Data Generating Process}
Let there be two mixture components, or groups, in the GMM: the privileged group $X^{priv}$, and the disadvantaged group $X^{dis}$, defined as follows:
$$ X^{priv} := X | X_{protected}=x^* \sim \mathcal{N}(\mu_{priv}, \sigma^2_{priv}) $$
$$ X^{dis} := X | X_{protected} \neq x^* \sim \mathcal{N}(\mu_{dis}, \sigma^2_{dis}) $$

Assume we get $n$ samples from the data generating process, and for each sample we get a scalar value $X$ sampled as described, along with a second binary-valued $X_{protected}$. 
The set-up is shown graphically in Figure \ref{fig:mean_estimation}.

\subsection{Defining Estimators}
Let's now fit different mean estimators under different modelling assumptions, namely, \iid and \ciid.

\subsubsection{Overall model (IID Assumption)}
\label{sec:iid-model}
In the \iid setting, we will fit a single model on the entire dataset. Here, the MLE is the sample average. 
\begin{equation}
        \hat{\mu}_{overall} =  \bar{X} = \frac{1}{n} \sum_{i=1}^{n} X^i 
\end{equation}
\subsubsection{Group-specific models (C-IID Assumption)}
\label{sec:ciid-model}
In the \ciid setting, we assume that data is a mixture coming from two different distributions, and we know apriori which of the two distribution samples are drawn from. To estimate the means of both distributions, we fit separate estimators for each mixture component. We assume that there are $n$ samples in total, and that $n_{priv}$ and $n_{dis}$ samples are drawn from $X^{priv}$ and $X^{dis}$, respectively.

\begin{equation}
        \hat{\mu}_{priv} =  \bar{X}^{priv} = \frac{1}{n} \sum_{i=1}^{n} X^i.\mathbbm{1} [X_{protected}= x^*]
\end{equation}
\begin{equation}
        \hat{\mu}_{dis} =  \bar{X}^{dis} = \frac{1}{n} \sum_{i=1}^{n} X^i.\mathbbm{1} [X_{protected}\neq x^*]
\end{equation}
\begin{equation}
        \hat{\mu}_{ciid} =  \hat{\mu}_{priv}.\mathbbm{1} [X_{protected}=x^*] + \hat{\mu}_{dis}.\mathbbm{1} [X_{protected}\neq x^*]
\end{equation}

We rewrite the overall model in terms of group-specific models. 
Denoting $p_{priv} = \frac{n_{priv}}{n}$ and $p_{dis} = \frac{n_{dis}}{n}$:

\begin{equation}
        \hat{\mu}_{overall} = \bar{X} = 
        p_{priv}.\hat{\mu}_{priv} + p_{dis}.\hat{\mu}_{dis}
\end{equation}

\subsubsection{Ensemble}
\label{sec:ensemble-model}
For comparison, we will also look at a simple ensemble model that averages the outputs of group-specific estimators.
\begin{equation}
        \hat{\mu}_{ensemble} = \frac{\hat{\mu}_{priv} + \hat{\mu}_{dis}}{2}
\end{equation}

\begin{table*}[t!]
    \centering
    \small 
        \caption{Demographic composition of the compas dataset (left to right): sex$\_$race$\_$priv$\_$priv is white women, sex$\_$race$\_$priv$\_$dis is non-white women, sex$\_$race$\_$dis$\_$priv is white men, sex$\_$race$\_$dis$\_$dis is non-white men, sex$\_$priv is women, sex$\_$dis is men, race$\_$priv is whites, and race$\_$dis is non-whites. Full is reported on the entire test set. Groups 1-3 are the groups assigned from unsupervised clustering.}
    \begin{tabular}{|c|c|c|c|c|c|c|c|c|}
        \hline
          & sex$\_$race$\_$priv$\_$priv& sex$\_$race$\_$priv$\_$dis& sex$\_$race$\_$dis$\_$priv & sex$\_$race$\_$dis$\_$dis& sex$\_$priv& sex$\_$dis& race$\_$priv& race$\_$dis   \\
         \hline
         Full & 0.083 & 0.105 &  0.321 & 0.491 & 0.188 & 0.812 & 0.404 &  0.596 \\
         \hline
         Group1 & 0.078 & 0.114 & 0.305 & 0.502 & 0.192 &  0.808 & 0.384 & 0.616 \\
         \hline
         Group2 & 0.132 & 0.105 & 0.351 & 0.412 & 0.237 &  0.763 &  0.483 &  0.517\\
         \hline
         Group3 & 0.027 & 0.059 & 0.144 & 0.769 & 0.087 &  0.913 &   0.172 &  0.828 \\
         \hline
    \end{tabular}
    \label{tab:compas-info}
\end{table*}

\subsection{Analysis of Trade-offs}
The bias-variance trade-offs of these estimators on different parts of the input data ($priv$ or $dis$ samples) 
is summarized in Table \ref{table:mean-results}. Here, $\Delta \mu$ = $|\mu_{priv} - \mu_{dis}|$ is the mean difference. See Appendix \ref{sec:appendix-theory} for a derivation.

Between all the models we analyze, the overall (\iid) and the \ciid fall on opposite ends of the spectrum. The \iid model has potentially unbounded bias: the bias grows as the mean difference grows. 
Further, if we overfit to the majority distribution using a single model, and observe only very few samples from the minority distribution during testing, we can overlook how biased the \iid model actually is, because the bias on samples from the majority class is low, and so the overall error will be low as well.

On the other hand, the \ciid model is unbiased by construction, but we pay for this in terms of large variance. However, we posit that in some contexts sacrificing estimator variance to gain unbiasedness on protected groups is a worthy trade-off.  We elaborate on this further, and will validate this conjecture experimentally in Section~\ref{sec:experiments}.

Considering the trade-offs made by the $dis$ model (trained on disprivileged group samples) in Table \ref{table:mean-results}, we build an intuition for when variance can be helpful, especially for ``fair'' classification. Over-fitting to the $dis$ group has two benefits: Firstly, we directly reduce the error on this demographic group.  This is morally desirable from a fairness perspective, since we want the error on the disprivileged group to be small, or at least comparable, to the error rates on privileged groups.
Secondly, the large variance of the conditional estimator can be a good thing: it allows us to still search for/land on ``other good values'' to estimate the mean of the second group (that it did not see during training). 
This is just not possible with the overall model because it biases towards the majority mixture component (\ie the privileged group) and also drastically reduces the variance, and any potentially corrective effect it could have had.

We see that the bias terms for all the models (whose bias is non-zero) depends on the mean difference $\Delta \mu$. This frames fair classification in the \ciid model as the task of domain adaptation, with the goal of fitting to the domain (\ie demographic group) that is most informative. We posit that this will be the group with the highest noise variance (\ie the disprivileged group). Indeed, as we will see in Section~\ref{sec:experiments}, 
the model trained on the disprivileged group will be unbiased for that group, and it will also have large enough variance to be able to adapt and make reasonably good predictions for the groups unseen during training. 

As we will further discuss in Section~\ref{sec:discussion}, randomness is also morally neutral form the philosophical perspective, and so the effects of large variance can be morally more acceptable than the effects of systematic skew in predictions.
Group-specific estimators are (by construction) unbiased on the population on which they were trained, and their large variance allows them to perform reasonably well for unseen groups. Hence, the net-effect of the large variance of conditional estimators can be \emph{fairness-enhancing}.

\section{Empirical Analysis}
\label{sec:experiments}
\subsection{Experimental Set-Up}
\label{sec:exp-setup}
\textbf{Datasets.} We used two fair-ml benchmark dataset in our evaluation, \href{https://github.com/zykls/folktables#1}{folktables}  and \href{https://github.com/propublica/compas-analysis}{COMPAS}.

\begin{table*}[t!]
    \centering
    \small 
    \begin{tabular}{|c|c|c|c|c|c|c|c|c|}
        \hline
          & sex$\_$race$\_$priv$\_$priv& sex$\_$race$\_$priv$\_$dis& sex$\_$race$\_$dis$\_$priv & sex$\_$race$\_$dis$\_$dis& sex$\_$priv& sex$\_$dis& race$\_$priv& race$\_$dis   \\
         \hline
         Full & 0.322 & 0.161 & 0.338 & 0.177 & 0.484 & 0.516 &  0.661 & 0.339 \\
         \hline
         Group1 & 0.317 & 0.201 & 0.295 & 0.186 & 0.519 &  0.481 & 0.612 &  0.388 \\
         \hline
         Group2 & 0.291 & 0.248 &  0.227 & 0.234 &  0.539 &  0.461 & 0.518 & 0.482 \\ 
         \hline
         Group3 & 0.320 &  0.118 & 0.369 & 0.192 &  0.438 &  0.561 & 0.689 & 0.311 \\
         \hline
         Group4 & 0.353 & 0.114 & 0.404 & 0.128 &  0.468 &  0.532 & 0.757 & 0.243 \\
         \hline
    \end{tabular}
    \caption{Demographic composition of the folktables dataset (left to right): sex$\_$race$\_$priv$\_$priv is white men, sex$\_$race$\_$priv$\_$dis is non-white men, sex$\_$race$\_$dis$\_$priv is white women, sex$\_$race$\_$dis$\_$dis is non-white women, sex$\_$priv is men, sex$\_$dis is women, race$\_$priv is whites, and race$\_$dis is non-whites. Full is reported on the entire test set. Groups 1-4 are the groups assigned from unsupervised clustering.}
    \label{tab:folk-info}
\end{table*}

Folktables~\citep{DBLP:conf/nips/DingHMS21} is constructed from census data from 50 US states for the years 2014-2018. We report results on the ACSEmployment task: a binary classification task of predicting whether an individual is employed. We report our results on data from Georgia from 2016, but observe consistent trends on different states and years. The dataset has 16 covariates, including age, schooling, and disability status, and contains about 200k samples. 

COMPAS~\citep{compas_propublica} is perhaps the most influential dataset in fair-ml, released for public use by ProPublica as part of their seminal report titled ``Machine Bias.'' COMPAS forms a binary classification task to predict violent recidivism. 
Covariates include sex, age, and information on prior criminal justice involvement. 
We use the version of COMPAS supported by \href{https://fairlearn.org/v0.4.6/auto_examples/plot_binary_classification_COMPAS.html}{fairlearn}. Fairlearn loads the dataset pre-split into training and test. We merge them into a single dataset and then perform different random splits. The full dataset has 5,278 samples. 

\begin{figure*}
    \centering
    \includegraphics[width=\linewidth]{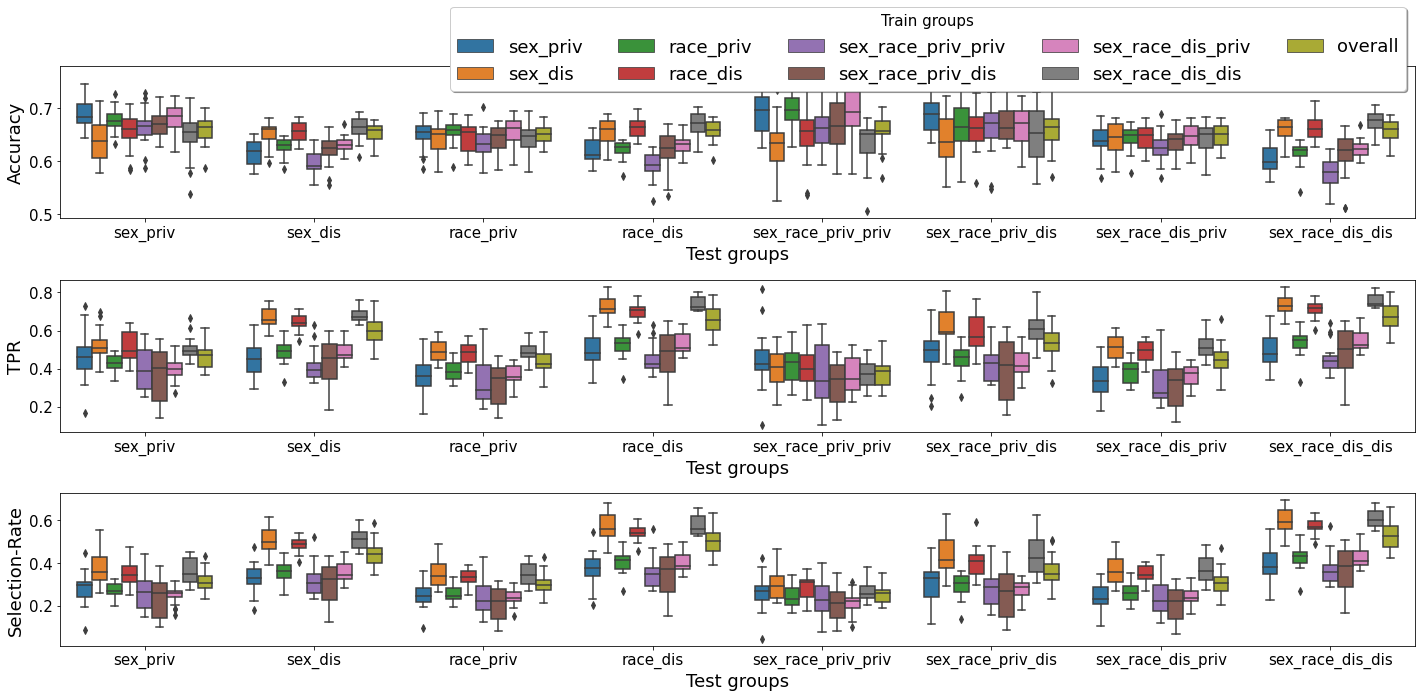}
    \caption{Conditioning on sensitive attributes, COMPAS: Test performance of different models broken down by test subgroup.}
    \label{fig:compas_groups}
\end{figure*}

\textbf{Protected Groups.} We define binary protected groups with respect to two features, sex and race. Males are the privileged ($priv$) group in folktables, while females are the privileged group in COMPAS. Whites are the privileged group in both folktables and COMPAS. We also look at intersectional groups constructed from sex and race: for example,  (male,white) is the priv$\_$priv group in folktables.
The proportion of demographic groups in folktables and COMPAS is reported in Tables \ref{tab:folk-info} and~ \ref{tab:compas-info}, respectively. 


\begin{figure*}[t!]
    \centering
    \includegraphics[width=\linewidth]{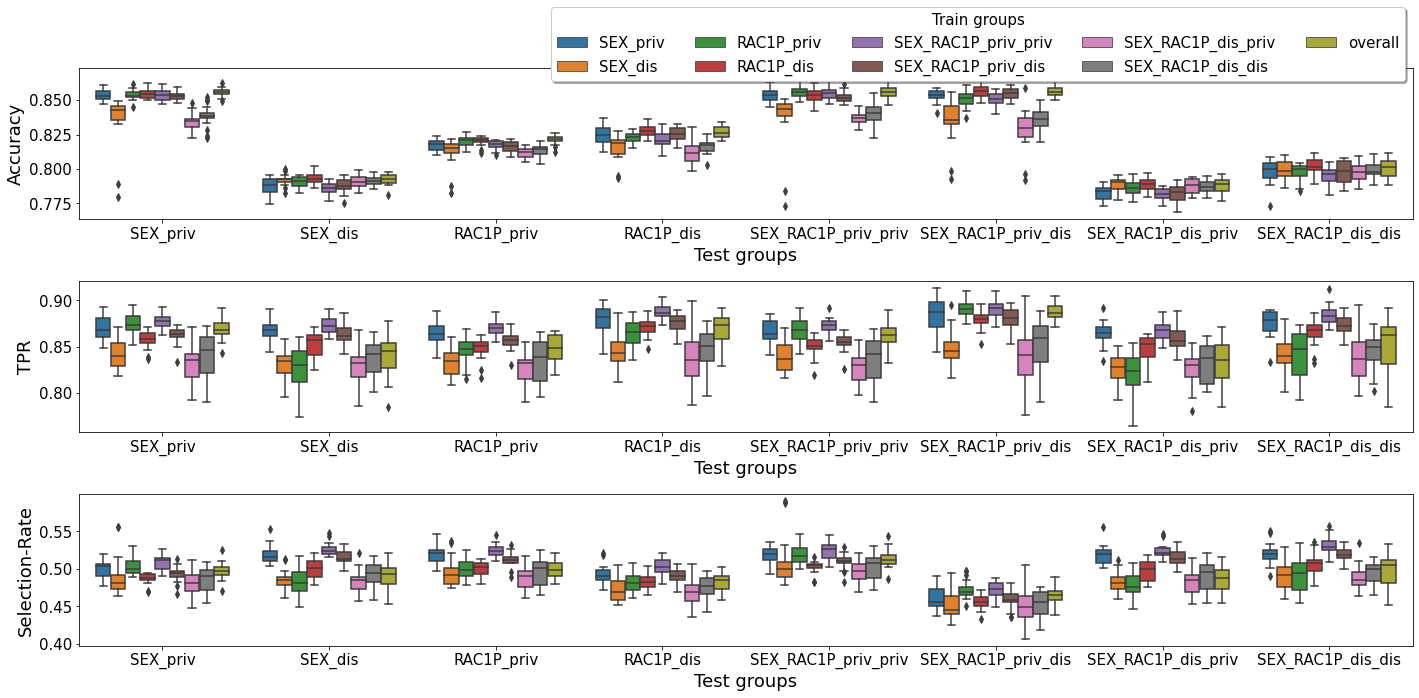}
        \vspace{-0.75cm}
    \caption{Conditioning on sensitive attributes, folktables: Test performance of different models broken down by test subgroup.}
    \label{fig:folk_groups}
\end{figure*}

\textbf{Training.} We fit the following models on a training set $\mathcal{D} (X^i,Y^i)$:
\begin{enumerate}
    \item Overall \iid model is trained on the entire  dataset, treating it as sampled from a single distribution. 

    \begin{equation} 
    \hat{f}_{overall}(\mathcal{D}) := argmin_{f \in \mathcal{F}} \mathbb{E}_{\mathcal{D}}[\ell (f(X),Y)]
    \end{equation}
    \begin{equation}
    \hat{y}_{iid}(X) = \hat{f}_{overall}(X_{relevant}, X_{protected})
    \end{equation}
    \item \Ciid models: We train two models, each on one subset of the data: the $priv$ model is fit on samples only from the privileged group. For example, the sex$\_$priv model in experiments on  folktables denotes the estimator that was fit on samples only from male respondents to the census. 
    Conversely, the $dis$ model only sees samples from the disadvantaged group during training. For example, the race$\_$dis model corresponds to the estimator fit on samples from non-white applicants. We apply the same procedure for intersectional groups. For example, the sex$\_$race$\_$dis$\_$dis model on COMPAS corresponds to the estimator fit on samples from male, non-white defendants.

\begin{equation}
    \hat{f}_{priv}(\mathcal{D}) := argmin_{f \in \mathcal{F}} \mathbb{E}_{\mathcal{D}}[\ell (f(X^{priv}),Y^{priv})] 
\end{equation}
\begin{equation}
    \hat{f}_{dis}(\mathcal{D}) := argmin_{f \in \mathcal{F}} \mathbb{E}_{\mathcal{D}}[\ell (f(X^{dis}),Y^{dis})] 
\end{equation}
\begin{equation}
\begin{aligned}
\hat{y}_{ciid}(X) = \hat{f}_{priv}(X_{relevant}).\mathbbm{1} [X_{protected}=x^*] \\ + \hat{f}_{dis}(X_{relevant}).\mathbbm{1} [X_{protected} \neq x^*]
\end{aligned}
\end{equation}
We will also look at these models in isolation, \ie if we applied a single conditional model to the entire population:
\begin{equation}
    \hat{y}_{priv}(X) =  \hat{f}_{priv}(X_{relevant})
\end{equation}
\begin{equation}
    \hat{y}_{dis}(X) =  \hat{f}_{dis}(X_{relevant})
\end{equation}
\end{enumerate}

\textbf{Model Selection and Testing.} 
\label{sec:model-training}
For each experiment, in one run we randomly split the data into train:test:validation (80:10:10), fit all the models described above, and report several predictive metrics: accuracy, true positive rate (TPR), and selection rate, on the test set. We chose these metrics because fairness measures are usually composed as differences and ratios of these performance metrics computed on different demographic groups in the test set~\cite{hardt_EOP2016, dwork_awareness, chouldechova_impossibility, Kleinberg_impossibility}. We use the validation set to tune hyperparameters once for each model type, for each dataset. Notably, the only difference between the  models we compare is the sub-population of the dataset that they see during training. Everything else, crucially including the model type (architecture and hyperparameters), is held fixed. For each experiment we conduct 18 runs, each with a different random split of the dataset. We use scikit-learn's implementations of different predictors. We experimented with different model types, namely LogisticRegression, MLPClassifier, DecisionTreeClassifier and KNeighborsClassifier and observed comparable results across all. We report results from the DecisionTreeClassifier here. 


\subsection{Experimental Results}
We run two sets of experiments: the first using training groups constructed on the basis of sensitive attributes (sex, race and sex$\_$race), and the second by clustering the samples into groups and constructing training groups based on cluster membership (blind to sensitive information). 

We present our results in two ways: we first report the performance of all models on different test subgroups. From a fairness perspective it is important that models perform comparably well (for our chosen metric) on all demographic groups in the test set. Next, we compare the performance of different models on the overall test set.

\subsubsection{Conditioning on sensitive attributes}
\label{sec:exp-sensitive}
The accuracy, TPR, and selection rate of different models on COMPAS and folktables is reported in Figures~\ref{fig:compas_groups} and~\ref{fig:folk_groups}. Colors denote different  models, and the $x$-axis reports performance broken down by test subgroup. We report several metrics on the overall test set in Figure \ref{fig:compas_all} (COMPAS) and \ref{fig:folk_all} (folktables). See Appendix~\ref{sec:appendix-sensitive} for results with other metrics. 


Our empirical results demonstrate that both overall performance, and performance on specific demographic groups, of group-specific models is comparable to performance of a single model trained on the entire training set on the overall test set (see Figure~\ref{fig:folk_groups} for folktables and Figure \ref{fig:compas_groups} for COMPAS). The results suggest that if samples are coming from two (or possibly more) distributions, there is a significant overlap between their supports --- models trained on only one subset of the data (\eg race$\_$priv,  trained only on white samples) perform competitively to models that see both groups during training (\eg the overall model), even on test groups that they never saw during training (\eg race$\_$dis, which corresponds to all non-white test samples). Furthermore, the conditional models are trained with at most half of the data compared to the overall model, and yet perform competitively --- a substantial computational gain. The proportions of samples from different groups is summarized in Tables~\ref{tab:folk-info} and~\ref{tab:compas-info}, and are indicative of the relative training set sizes of different conditional estimators. 

From a fairness perspective, we desire that the model perform equally well on different demographic subgroups in the data, and the disparity in performance metrics (such as accuracy, FPR, selection rate, etc) are quantified as measures of model unfairness. Conditional estimators trained only on samples from the \emph{dis} group are more ``fair'' than \iid estimators, by construction: the race$\_$dis, sex$\_$dis and sex$\_$race$\_$dis$\_$dis are top performing models on their respective demographic groups in the test set as seen in Figure \ref{fig:folk_groups} (folktables) and Figure \ref{fig:compas_groups} (COMPAS). Additionally, these models are competitive even on \emph{priv} test groups that they did not see during training, and perform only marginally worse than the overall model. Hence, the disparity in the performance of these conditional models on \emph{dis} and \emph{priv} groups is smaller than that of the overall model, and we get the desired improvement in \emph{fairness}.

\subsubsection{Conditioning on cluster assignment (blind to protected group membership)}
\label{sec:exp-blind}
We also evaluate the efficacy of a \ciid model that is blind to protected group membership. In order to do so, we first cluster all the data points into a suitable number of groups and then train models after conditioning on cluster membership. For example, the model for Group1 is the estimator that only saw samples from the first cluster during training. 
Notably, the choice of the number of clusters is a hyperparameter of this blind conditional model. We find that choosing 3 and 4 clusters for COMPAS and folktables, respectively, creates clusters of a reasonable size and with reasonable coverage of protected groups. The proportion of protected groups in each cluster is reported in Tables~\ref{tab:compas-info} (COMPAS) and~\ref{tab:folk-info} (folktables). For COMPAS Group1, Group2 and Group3 constitute 56$\%$,  33.6$\%$ and 10.4$\%$ of the training dataset, respectively. For folktables, Groups 1-4 constitute 20.1$\%$, 22.1$\%$, 23.3$\%$m and  34.5$\%$ of the training data, respectively.

\begin{figure*}
    \centering
    \includegraphics[width=\linewidth]{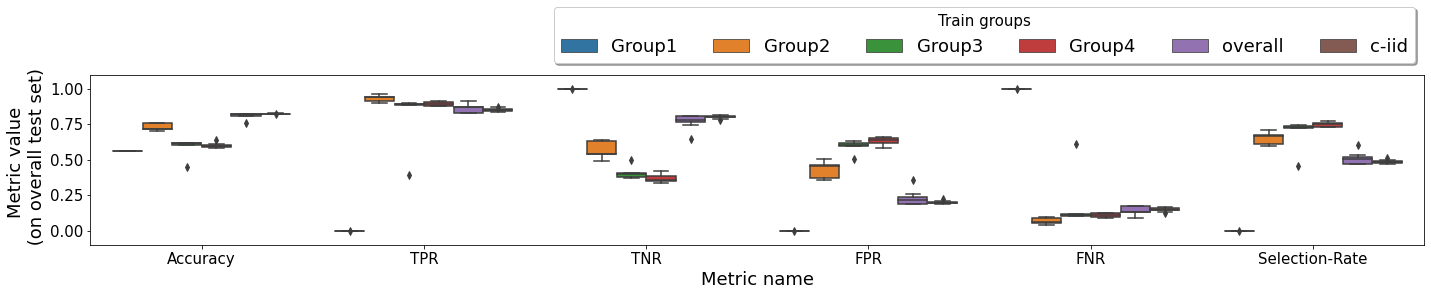}
        \vspace{-0.75cm}
    \caption{Conditioning on cluster membership (blind), folktables: Test performance of different models on the overall test set.}
    \label{fig:folk_clusters_all}
\end{figure*}
\begin{figure*}
    \centering
    \includegraphics[width=\linewidth]{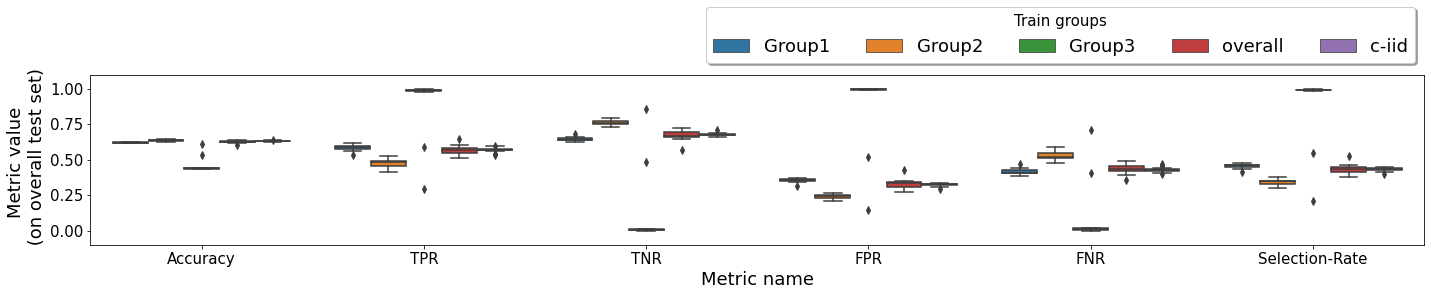}
    \vspace{-0.75cm}
    \caption{Conditioning on cluster membership (blind), COMPAS: Test performance of different models on the overall test set.}
    \label{fig:compas_clusters_all}
\end{figure*}
As before, we report several metrics on the overall test set in Figure \ref{fig:folk_clusters_all} (folktables) and Figure \ref{fig:compas_clusters_all} (COMPAS). We report additional metrics, as well as a breakdown by demographic groups in the test set, in Appendix~\ref{sec:appendix-blind}.

The results of this experiment demonstrate the efficacy of a race-blind and sex-blind classifier in the \ciid setting. We see a slightly different mechanism that is leading to gains in performance, however: unsupervised clustering before conditioning allows us to identify and isolate spurious samples. The Group1 estimator (in green) in Figure~\ref{fig:compas_clusters_groups} (COMPAS) and the Group1 estimator (in blue) in Figure~\ref{fig:folk_clusters_groups} (folktables) are clearly trained on spurious samples. In practice, \ciid allows us to identify and thereby isolate such spurious samples in the training set.

\section{Discussion}
\label{sec:discussion}

\paragraph{Worldview.} 
Our work aligns with the 
``We Are All Equal'' (WAE) worldview of~\citet{friedler_impossibility} that posits that people's qualifications are indistinguishable in the construct space, but, due to systematic skew (the effect of discrimination), they are no longer identically distributed in the observed space. From a philosophical perspective, the WAE worldview necessitates modelling beyond the \iid setting. 

\paragraph{Desert vs. Privilege.} From a moral perspective, it is desirable to use desert-based decision-making procedures in critical contexts such as criminal justice and employment. A desert-based procedure operates on whether the applicant deserves the outcome, whether good (positive employment status) or bad (high risk of recidivism). Our compelling empirical results on COMPAS and folktables in Section~\ref{sec:experiments} are an indication that real-world decision-making is influenced more by privilege than by desert. Here, model unfairness is evidence of the massive weight of privilege: our datasets are records of past decision-making. If this was purely desert-based, then seeing samples from socially privileged groups (such as sex$\_$priv and race$\_$priv) would not inherently make the procedure perform worse on samples from socially disadvantaged groups. 

\paragraph{Rawls's Original Position.}
Philosophically, our porposed procedure resembles John Rawls's Original Position under a Veil of Ignorance. Rawls posits that people would select rules of social cooperation under a veil of ignorance, \ie without knowing where they are going to end up on the social ladder in a way that agrees with his principles of justice~\cite{Rawls1971Justice}. The intuition is that, if nobody knows where they are going to end up, and could possibly end up on the bottom-most rung of the ladder, then everyone will set up society in a just manner, keeping in mind the position of the worst-off group. In our set-up, the most socially disadvantaged group is the intersectionally disadvantaged one. Hence, a procedure that trains an estimator on samples from that group (the sex$\_$race$\_$dis$\_$dis estimator in Section~\ref{sec:exp-sensitive}), and then applies it to everyone, irrespective of their morally irrelevant characteristics (sensitive features), formalizes the notion of fairness that John Rawls' posits in this broad theory of justice~\cite{Rawls1971Justice}.

\paragraph{Procedural Fairness.}
The \ciid model uses group-specific estimators during testing. For example, the race$\_$priv model is used to predict outcomes for white applicants, whereas the race$\_$dis model is applied to non-white applicants. While this gives us the improved predictive and computational gains we desired, from a legal standpoint, the \ciid model is procedurally unfair because it is unfair to apply different decision-making procedures to applicants based on their sensitive features.

That being said, we also provide two conditional formulations that satisfy procedural fairness: the first is simply to use one of the conditional models on the entire population. For example, the procedure described under ``Rawls's Original Position'' uses only the sex$\_$race$\_$dis$\_$dis to make predictions. This is no longer a violation of procedural fairness. Although the estimator only saw samples from one demographic group during training, it is the very same estimator being used to decide outcomes for all candidates, irrespective of their sensitive attributes, and hence this is a fair procedure. The blind classifier described in Section~\ref{sec:exp-blind} is another way to prevent procedural unfairness in the \ciid setting. 



\section{Conclusions, Limitations and future work}
In this paper we challenged the suitability of the dominant \iid setting for fair decision-making. Through our theoretical analysis in the Gaussian mixture model, and empirical evaluation on benchmark datasets, we hope to have demonstrated the suitability of an alternate formulation: the \ciid setting. We also instantiated the proposed \ciid model with two different procedures that improve ``fairness'' by conditioning out undesirable effects: first, by conditioning directly on sensitive attributes, and, second, by clustering samples into groups and conditioning on cluster membership. 

Our work opens many avenues for exciting future work.  One of these is that, in social applications we have the luxury of modelling in the low-dimensional region of feature space: There are 17 categorical features in folktables, and on the order of 200k samples. In COMPAS, there are 13 features, and on the order of 5k samples. The \ciid setting is suitable for social domains such as the tasks evaluated here because we are in the low-dimensional regime. There is interesting future work to be done to investigate the efficacy of the \ciid model beyond tabular data, for example, 
on image data with thousands of pixels, especially in medical contexts, where the number of features can be larger than the number of labelled samples.

Furthermore, in our experiments we showed that most of the conditional models are competitive with models that have seen the full dataset. Notably, even intersectionally group-specific models like the 
sex$\_$race$\_$dis$\_$dis estimator and sex$\_$race$\_$dis$\_$priv estimator --- which were trained only on 17.7$\%$ and 33.8$\%$ of folktables, and 49.1$\%$ 32.1$\%$ of COMPAS, respectively --- are highly competitive.  Interestingly, these are also the test groups that most models perform the poorest on! This is extremely telling about the informativeness of samples from the intersectionally disadvantaged groups. The deceptively simple point we demonstrated is that samples from marginalized groups are more likely to form the predictive margin. There is interesting future work to be done to connect this bias-variance analysis with margin analysis in the \ciid setting. 

\bibliographystyle{ACM-Reference-Format}
\bibliography{sample-base}

\appendix
\section{Bias-Variance Analysis in the Gaussian Mixture Model}
\label{sec:appendix-theory}

\subsubsection{Overall Model}
Let's begin by computing the bias on $X^{priv}$ samples:
$$ \mathbb{E}[\hat{\mu}_{overall} - \mu_{priv}] = 
\mathbb{E}[p_{priv}.\hat{\mu}_{priv} + p_{dis}.\hat{\mu}_{dis} - \mu_{priv}] $$
$$ =\mathbb{E}[p_{priv}.\hat{\mu}_{priv} + p_{dis}.\hat{\mu}_{dis} - p_{priv}.\mu_{priv} - p_{dis}\mu_{priv}] $$
\begin{equation}
        p_{priv}\mathbb{E}[\hat{\mu}_{priv} - \mu_{priv}] + p_{dis}\mathbb{E}[\hat{\mu}_{dis} - \mu_{priv}]
\end{equation}

The first term goes to zero since $\hat{\mu}_{priv}$ is an unbiased estimator of $\mu_{priv}$. We can approximate the second term using the mean difference $\Delta \mu$, defined as follows (also shown pictorically in Figure \ref{fig:mean_estimation}):
\begin{equation}
        \Delta \mu = |\mu_{priv} - \mu_{dis}|
\end{equation}

Plugging this into (7), we get:
$$ \mathbb{E}[\hat{\mu}_{overall} - \mu_{priv}] = p_{dis}\mathbb{E}[\hat{\mu}_{dis} - \mu_{priv}] $$
\begin{equation}
\approx  p_{dis}\mathbb{E}[\mu_{dis} - \mu_{priv}] = p_{dis}.\mathbb{E}[\Delta \mu]
\end{equation}

Similarly, we can derive the expression for bias on $X^{dis}$ samples as:
\begin{equation}
\mathbb{E}[\hat{\mu}_{overall} - \mu_{dis}] = p_{priv}\mathbb{E}[\hat{\mu}_{priv} - \mu_{dis}] \approx p_{priv}.\mathbb{E}[\Delta \mu]
\end{equation}

This is a very intuitive result: we are paying for mis-specifying a single model instead of group-specific ones. The bias term is simply the product of the mean difference (how ``far" the two distributions are from each other) and the proportion of samples from the ``wrong" group (how many "bad" samples we used in our estimation).

Let's look at the variance (which does not depend on the true value) of this estimator next:
$$ \text{Var}(\hat{\mu_{overall}}) = \text{Var}(p_{priv}.\hat{\mu}_{priv} + p_{dis}.\hat{\mu}_{dis}) $$
\begin{equation}
= p_{priv}^2.\text{Var}(\hat{\mu}_{priv}) + p_{dis}^2.\text{Var}(\hat{\mu}_{dis}) + 2.p_{priv}^2.p_{dis}^2.\mathbb{E}[[\hat{\mu}_{priv} - \mathbb{E}\hat{\mu}_{priv}][\hat{\mu}_{dis} - \mathbb{E}\hat{\mu}_{dis}]]
\end{equation}

The third term goes to zero since $\hat{\mu}_{priv}$ and $\hat{\mu}_{dis}$ are unbiased. Let us write expressions for the variance of these two estimators now. Recall that these are both simply the MLE of their respective distribution, so we know their variance scales in the order of 1/n:
\begin{equation}
  \text{Var}(\hat{\mu}_{priv}) = \frac{\sigma^2_{priv}}{n_{priv}}  
\end{equation}

\begin{equation}
  \text{Var}(\hat{\mu}_{dis}) = \frac{\sigma^2_{dis}}{n_{dis}}  
\end{equation}

Plugging this back into (11):
$$ \text{Var}(\hat{\mu_{overall}}) = p_{priv}^2.\frac{\sigma^2_{priv}}{n_{priv}} + p_{dis}^2.\frac{\sigma^2_{dis}}{n_{dis}} 
$$
\begin{equation}
    = \frac{1}{n} [\frac{n_{priv}}{n}.\sigma^2_{priv} + \frac{n_{dis}}{n}.\sigma^2_{dis}]
\end{equation} 

\subsubsection{Conditional Model}
The conditional model defined in section \ref{sec:ciid-model} is by construction unbiased: we have fit unbiased models to individual mixture components, and will use a-priori information about mixture membership ($X_{protected}$) to decide which estimator to use.

Let's look at it's variance: the conditional estimator uses $\hat{\mu}_{priv}$ to estimate the mean when samples come from $X^{priv}$, and $\hat{\mu}_{dis}$ when samples come from $X^{dis}$. So, the conditional model has a variance of $\sigma_{priv}^2/n_{priv}$ and $\sigma_{dis}^2/n_{dis}$ on samples from $X^{priv}$ and $X^{dis}$ respectively. 

Although the difference between the i.i.d and conditional-i.i.d setting might seem subtle, we can already start seeing how different the resulting procedures are: in the i.i.d case we paid heavily in terms of bias: if distributions are very far apart ($\Delta \mu$ is large), and/or if samples are highly unequally drawn from mixture components ($p_priv$ or $p_dis$ is close to 1, and the other is close to 0) then errors are large and unequally distributed: we do well on the distribution whose mean is closer to the estimated value, and do very poorly on the other mixture component. In the c-i.i.d setting we have the opposite problem: we are able to construct unbiased estimators for both mixture components. However in doing so, the variance of our estimator also becomes conditional on group membership --- we pay in the order of $\sigma^2_{group}/n_{group}$. So, for groups/mixture components for which we see very little data, or very noisy data, we have large variance. 

\subsubsection{Ensemble Model}
Recall the ensemble model defined in \ref{sec:ensemble-model}. This is a symmetric model, and so the bias is the same on samples from the privileged group and the disadvantaged group. Writing it for the priv group:
\begin{equation}
 \mathbb{E}[\hat{\mu}_{ensemble} - \mu_{priv}] = \mathbb{E}[\frac{\hat{\mu}_{priv} + \hat{\mu}_{dis}}{2} - \mu_{priv}] = \frac{1}{2}\mathbb{E}[\hat{\mu}_{dis} - \mu_{priv}] \approx \frac{1}{2} \mathbb{E}[\Delta \mu]   
\end{equation}

And, its variance is:
\begin{equation}
 \text{Var}[\hat{\mu_{ensemble}}] = \text{Var}[\frac{\hat{\mu}_{priv} + \hat{\mu}_{dis}}{2}] = \frac{1}{4} [\text{Var}(\hat{\mu}_{priv}) + \text{Var}(\hat{\mu}_{dis})] = \frac{1}{4} (\frac{\sigma^2_{priv}}{n_{priv}} + \frac{\sigma^2_{dis}}{n_{dis}})
\end{equation}

\subsubsection{Dis Model}
Lastly, let's look at how we would perform if we only used the model trained on $X^{dis}$ samples, ie. $\hat{\mu}_{dis}$. 
We know that $\hat{\mu}_{dis}$ is an unbiased estimator of ${\mu}_{dis}$, so the bias on samples from $X^{dis}$ is zero. Let's look at the bias on samples from $X^{priv}$:
\begin{equation}
    \mathbb{E}[\hat{\mu}_{dis} - \mu_{priv}] \approx \mathbb{E}[\mu_{dis} - \mu_{priv}] = \mathbb{E}[\Delta \mu].
\end{equation}

We already saw in (13) that the variance of this estimator is $\sigma^2_{dis}/n_{dis}$

\section{Conditioning on Sensitive Attributes}
\label{sec:appendix-sensitive}
\begin{figure*}
    \centering
    \includegraphics[width=\linewidth]{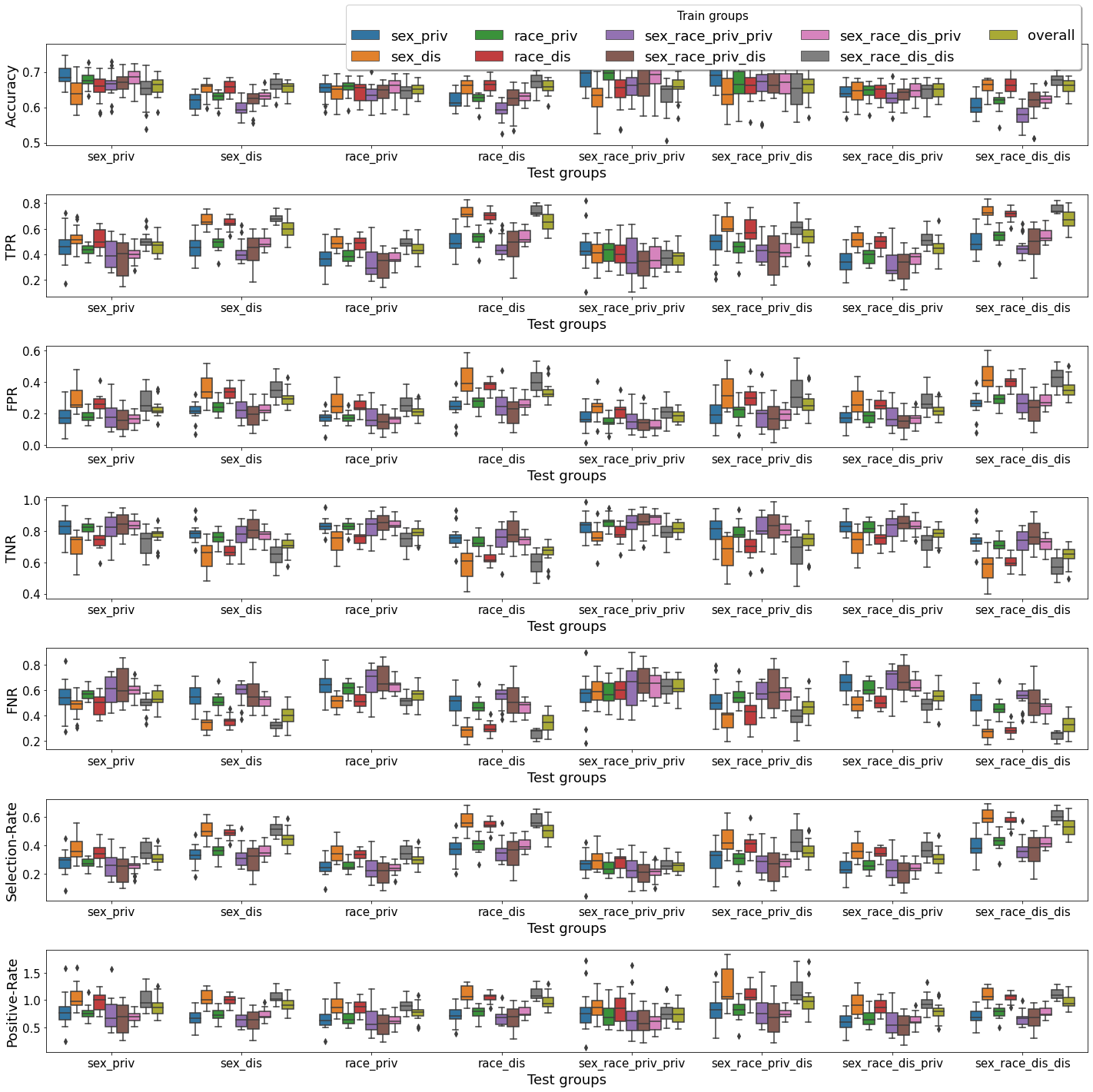}
    \caption{Conditioning on sensitive attributes, compas: Test performance of different models broken down by test subgroup.}
    \label{fig:compas_clusters_groups_all_metrics}
\end{figure*}

\begin{figure*}
    \centering
    \includegraphics[width=\linewidth]{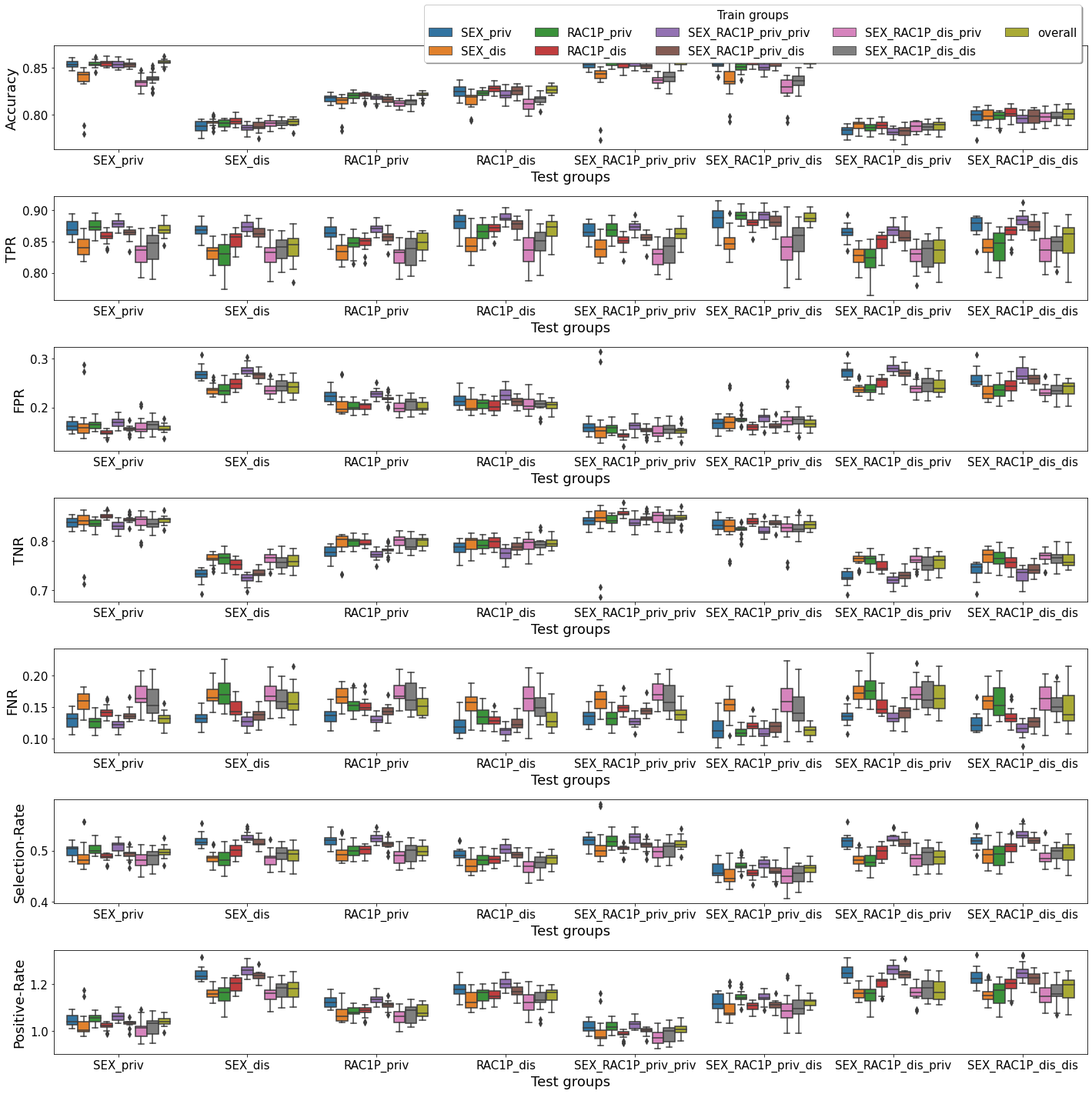}
    \caption{Conditioning on sensitive attributes, folktables: Test performance of different models broken down by test subgroup.}
    \label{fig:folk_clusters_groups_all_metrics}
\end{figure*}

Several performance metrics (accuracy, TPR, FPR, FNR, TNR, selection rate and positive rate) of different models on COMPAS and folktables is reported in Figures~\ref{fig:compas_clusters_groups_all_metrics} and~\ref{fig:folk_clusters_groups_all_metrics}. Colors denote different  models, and the $x$-axis reports performance broken down by test subgroup. Train groups are constructed on the basis of sensitive attributes. 

\section{Conditioning on Cluster Membership (blind)}
\label{sec:appendix-blind}
\begin{figure*}
    \centering
    \includegraphics[width=\linewidth]{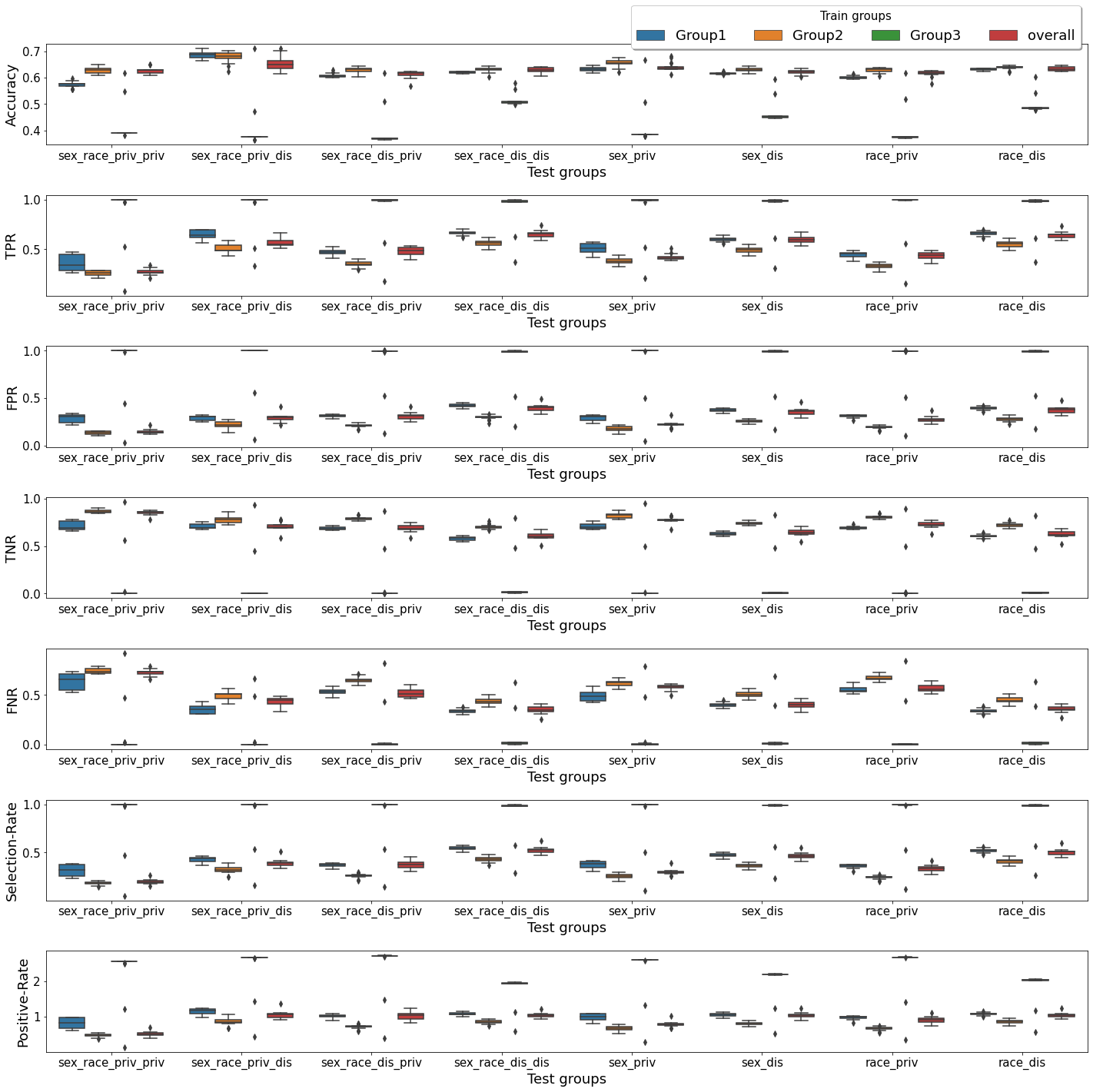}
    \caption{Conditioning on cluster membership (blind), compas: Test performance of different models broken down by test subgroup.}
    \label{fig:compas_clusters_groups}
\end{figure*}

\begin{figure*}
    \centering
    \includegraphics[width=\linewidth]{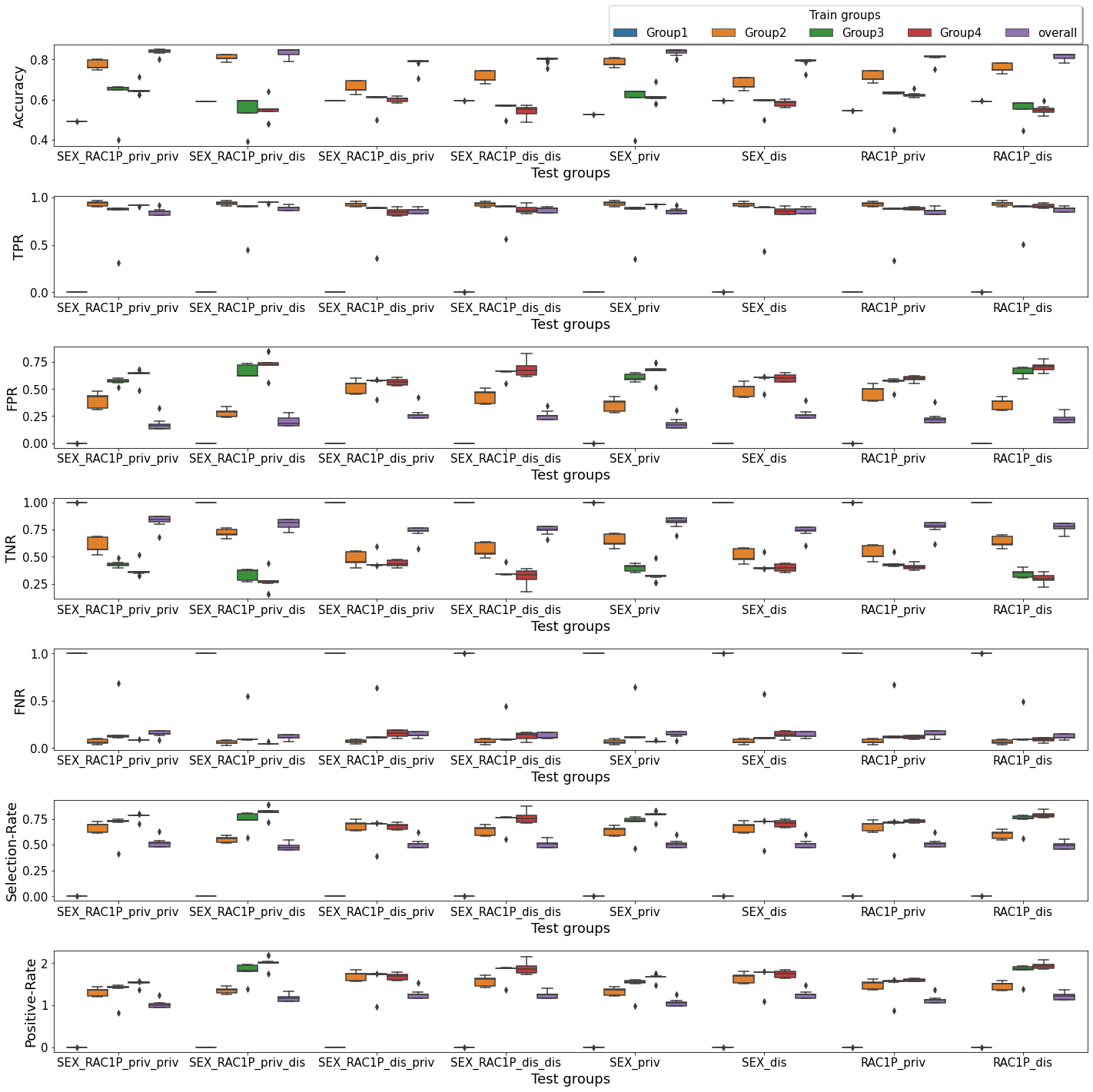}
    \caption{Conditioning on cluster membership (blind), folktables: Test performance of different models broken down by test subgroup.}
    \label{fig:folk_clusters_groups}
\end{figure*}

Several performance metrics (accuracy, TPR, FPR, FNR, TNR, selection rate and positive rate) of different models on COMPAS and folktables is reported in Figures~\ref{fig:compas_clusters_groups} and~\ref{fig:folk_clusters_groups}. Colors denote different  models, and the $x$-axis reports performance broken down by test subgroup. Train groups are constructed on the basis of cluster assignment. 

\end{document}